\documentclass[number,preprint,review,12pt]{elsarticle}

\usepackage{amssymb}
\usepackage{graphicx}
\usepackage{amsmath}
\usepackage{amssymb}
\usepackage{booktabs}
\usepackage{multirow}
\usepackage{xcolor}
\usepackage[authoryear]{}

\definecolor{mygray}{RGB}{128,128,128}
\usepackage[pagebackref,breaklinks,colorlinks]{hyperref}

\usepackage[capitalize]{cleveref}
\crefname{section}{Sec.}{Secs.}
\Crefname{section}{Section}{Sections}
\Crefname{table}{Table}{Tables}
\crefname{table}{Tab.}{Tabs.}

\journal{Pattern Recognition}

\begin{document}

\begin{frontmatter}

\title{Enhancing Weakly Supervised Semantic Segmentation with Multi-modal Foundation Models: An End-to-End Approach}

\cortext[cor1]{Corresponding author}

\author[1]{Elham Ravanbakhsh }\ead{eravan1@lsu.edu}
\author[2]{Cheng Niu  }\ead{chengniu@tamu.edu}
\author[2]{Yongqing Liang  }\ead{lyq@tamu.edu}
\author[1]{J. Ramanujam  }\ead{jxr@cct.lsu.edu}
\author[2,3]{Xin Li \corref{cor1} } \ead{xinli@tamu.edu}

\affiliation[1]{organization={Louisiana State University},
            addressline={Baton Rouge, LA 70803},
            country={USA}}

\affiliation[2]{organization={Department of Computer Science $\&$ Engineering, Texas A$\&$M University},
            addressline={College Station, TX 77843},
            country={USA}}

\affiliation[3]{organization={Section of Visual Computing and Interactive Media, Texas A$\&$M University}, addressline={College Station, TX 77843}, country={USA}}

\begin{abstract}
Semantic segmentation is a core computer vision problem, but the high costs of data annotation have hindered its wide application. Weakly-Supervised Semantic Segmentation (WSSS) offers a cost-efficient workaround to extensive labeling in comparison to fully-supervised methods by using partial or incomplete labels. Existing WSSS methods have difficulties in learning the boundaries of objects leading to poor segmentation results. 
We propose a novel and effective framework that addresses these issues by leveraging visual foundation models inside the bounding box. Adopting a two-stage WSSS framework, our proposed network consists of a pseudo-label generation module and a segmentation module. The first stage leverages Segment Anything Model (SAM) to generate high-quality pseudo-labels. To alleviate the problem of delineating precise boundaries, we adopt SAM inside the bounding box with the help of another pre-trained foundation model (e.g., Grounding-DINO). Furthermore, we eliminate the necessity of using the supervision of image labels, by employing CLIP in classification. Then in the second stage, the generated high-quality pseudo-labels are used to train an off-the-shelf segmenter that achieves the state-of-the-art performance on PASCAL VOC 2012 and MS COCO 2014.
\end{abstract}

\begin{highlights}
\item Novel framework for weakly supervised semantic segmentation
\item Integration of SAM inside the bounding box for fine-grained pseudo-label generation
\item Elimination of image-level labels using CLIP
\item State-of-the-art performance on PASCAL VOC 2012 and MS COCO 2014
\end{highlights}


\end{frontmatter}


\section{Introduction}
\label{sec:intro}
Semantic segmentation categorizes and labels pixels within an image into a class or object. It is a core computer vision task for many downstream applications like autonomous driving, medical image analysis and mobile robots. However, semantic segmentation's application in many practical tasks has been impeded by the need for costly pixel-level annotations to support supervised training. For example, it has been reported that annotating masks for ~164K images in the MS COCO dataset (which contains only 80 classes) consumed more than 28K human hours of annotation time \cite{wang2023cut}. As a result, the time and resource-intensive nature of manual pixel-level annotation limits the feasibility of supervised semantic segmentation in various settings. 

Weakly Supervised Semantic Segmentation (WSSS) emerges as a strategy aimed at eliminating the expense of annotation. It attempts to study a segmentation network that groups an image into coherent regions corresponding to different object categories or semantic entities with weaker forms of supervision than pixel-wise labeling, including image labels \cite{lee2021anti, wei2017object}, point labels \cite{khoreva2017simple}, and boxes \cite{lee2021bbam, oh2021background}. 
However, obtaining labels, even image labels which are the weaker form of supervision, can be challenging. This is particularly true in scenarios such as pictures captured in the wild, frames extracted from videos or movies, or recordings from autonomous vehicles, where labeling is not straightforward. 

Most existing WSSS approaches rely on a Class Activation Map (CAM) for obtaining location information. It usually consists of two stages: pseudo-label generation module and segmentation module. In the pseudo-label generation module, initially, a classifier is trained using image labels. This model learns to recognize different classes within the images but lacks specific details regarding object boundaries or precise locations. Following the model training, Class Activation Maps (CAMs) are produced. CAMs provide a basic indication of where different classes might exist in the images. They're derived from analyzing the intermediate feature maps created during classification to highlight regions associated with specific classes. Subsequently, the initial CAMs are refined and converted into more detailed pseudo-labels. This refinement involves techniques like pixel affinity-based methods \cite{li2023transcam}, bootstrapping \cite{yin2022transfgu} or saliency guidance \cite{lee2021railroad}, aiming to convert the coarse CAMs into finer, pixel-level annotations. This step enhances the delineation of object boundaries and finer image details. Lastly, in the segmentation module, a semantic segmentation model is trained using the refined pseudo-labels. This model learns to perform detailed pixel-level segmentation, effectively identifying and outlining specific objects or regions within the images based on the refined annotations.

\begin{figure}
\centering
\includegraphics[width=13.8cm,height=10cm,keepaspectratio]{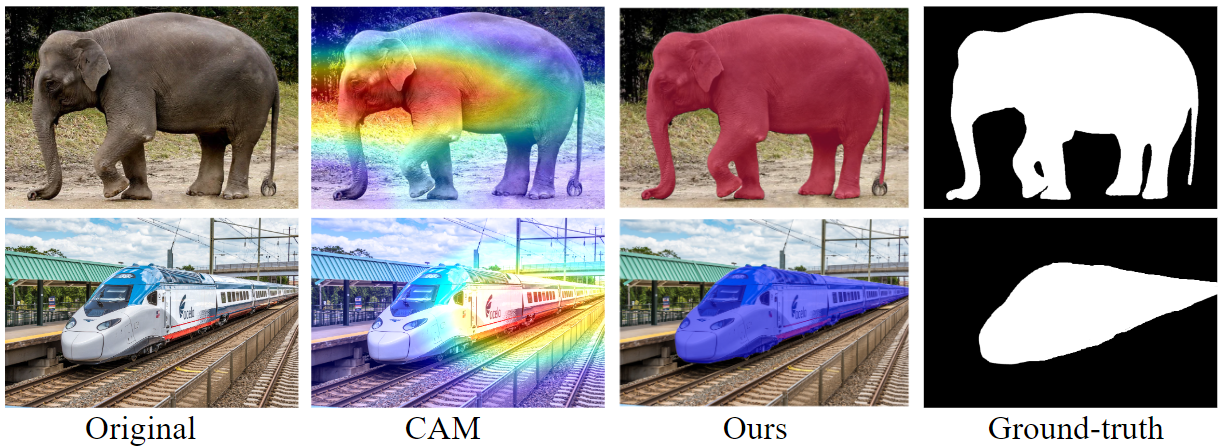}
\caption{\label{fig:issues} Illustration of the partial and false activation issues of CAMs on PASCAL VOC 2012 train set, our result (based on segmentation inside the bounding box), and the ground-truth. Row 1 presents the partial activation of CAMs while Row 2 demonstartes the false activation. }
\end{figure}

Clearly, the performance of WSSS is significantly dependent on the accuracy of the pseudo labels. However, the quality of pseudo-labels generated using CAM-based methods has not yet attained the standard of manually annotated masks. This discrepancy primarily stems from inaccuracies in delineating object boundaries from either partial activation \cite{ahn2018learning} or false activation \cite{xie2022clims}. Partial activation means that the classifier mainly emphasizes the most discriminative part of an object rather than the entire object area. For example, a classifier may usually activate parts mostly related to the head, face, and tail area of an elephant rather than its whole body as shown in Figure \ref{fig:issues}, row 1. False activation means sometimes CAMs encompass the background area around the object. For example, for recognizing a train, the railroad area may be also activated as shown in Figure
\ref{fig:issues}, row 2.

In this paper, we address these two issues by replacing CAM-based pseudo-labels with  detailed fine-grained labels derived from an advanced segmentation foundation model inside the bounding box. We leverage foundation models due to their strong capability of generating fine-grained, accurate masks of objects or parts to avoid partial activation. We also adopt the strategy of \textbf{segmentation inside the bounding box} to fix the issue with false activation. In our architecture, we specifically use Segment Anything Model (SAM) \cite{kirillov2023segment} due to its power in segmenting objects. 
Some existing foundation model-assisted WSSS methods such as \cite{chen2023weakly, yang2024foundation}, leverage SAM in enhancing their segmentation seeds generated by CAMs. They assign class labels to SAM-based masks by computing the overlap ratio to CAM-based pseudo labels. However, the CAM-based pseudo labels are prone to partial and false activation which may lead to false selection of masks. While by using SAM partial activation could be remedied to some extent, false activation issues still persist with their approach. We propose searching for pseudo-labels inside the bounding box. This will guides SAM to better segment each specific object by ensuring it is both class-aware and location-aware, and hence, effectively reduce false activation. To mitigate partial activation, we introduce a hierarchical grouping scheme for masks within the bounding box, prioritizing the recognition of entire objects over their parts and subparts. We surpass existing approaches by a significant margin on both PASCAL VOC and MS COCO.  In our architecture, we leverage Grounding-DINO \cite{liu2023grounding} for finding the bounding box containing the location of objects. Then we use SAM to predict pixel-level annotations inside the box. 
Furthermore, to eliminate the necessity of using image labels, we employ contrastive language-image pre-training (CLIP) \cite{radford2021learning} to predict labels. Unlike previous approaches, our proposed pipeline empowers us to generate class-aware pseudo-labels in an end-to-end manner.

We demonstrate that the quality of the SAM-based pseudo-labels surpasses the CAM-based approaches and even outperforms recent methodologies that integrate both coarse class locations from CAMs and object boundary information from SAM. 

Our proposed model alleviates the most important issues of CAM-based methods and achieves state-of-the-art performance. In summary, Our main contributions are summarized as follows:
\begin{itemize}
  \item We introduce a new framework for weakly supervised semantic segmentation that alleviates the issues with CAM-based methods by using SAM inside the bounding box. Our experiments demonstrate that we outperform state-of-the-art WSSS methods in generating pseudo labels. With these pseudo labels, our trained segmenter outperforms state-of-the-art WSSS models on commonly adopted benchmarks such as Pascal VOC 2012 and MS COCO 2014.
  \item Unlike other existing WSSS methods that require image labels (either class or bounding-box labels), we propose to use CLIP in WSSS setting to eliminate the necessity of using any image label for training, which could be unavailable or expensive to get in practical scenarios. 
\end{itemize}

\section{Related Work}
WSSS aims at learning pixel-level representations for semantic segmentation in the absence of pixel annotations with the help of any weaker form of annotations. Existing WSSS methods can be broadly categorized into one-stage and two-stage techniques. One-stage methods \cite{araslanov2020single, zhang2022end}, involve end-to-end training of a segmentation network using image-level labels. These approaches learn by enabling the network to autonomously extract intricate features and relevant information from the input data without needing to explicitly define intermediate representations. Conversely, two-stage methods \cite{yuan2023multi, yi2022weakly, kho2022exploiting, luo2021weakly} initially generate segmentation seeds in stage one and then utilize them as pseudo labels to train an off-the-shelf segmentation network. The effectiveness of WSSS is heavily dependent on the accuracy and completeness of pseudo-labels. Several techniques are proposed to generate and improve the quality of pseudo labels.

\subsection{CAM-based Pseudo-labels }
Most existing approaches train a classification network to generate class activation maps (CAM) \cite{zhou2016learning} serving as the initial pseudo labels. However, CAMs suffer from partial and false activation by activating only the most discriminative parts of a visual object or the background area, rather than the whole object area \cite{chen2023segment}. The reason is that the network's training process is guided by the classification loss, which primarily focuses on distinguishing between different classes. As the goal is often to identify the most discriminative regions for improved distinctiveness, the networks may not necessarily need to discover the entire object which leads to poor-quality pseudo-labels. 

To improve the quality of CAM-based pseudo-labels, several methods train a classification network with auxiliary tasks to guide the model toward the discovery of more object regions. Some approaches intentionally conceal or erase specific regions of an object, compelling models to explore more diverse parts \cite{kumar2017hide, wei2017object}. However, these methods either randomly hide fixed-size patches or necessitate repetitive model training and response aggregation steps \cite{chang2020weakly}. Other works adopt an adversarial erasing strategy \cite{zhang2018adversarial, li2018tell} to include more regions but they suffer from false activation, wrongly activating the background around the object. Some other approaches \cite{hou2018self, lee2019ficklenet} adopt different techniques including self-erasing strategy or stochastic feature selection to alleviate this issue but their performance is limited in obtaining high-quality labels. 

Different from previous approaches, some others deploy contrastive learning \cite{bachman2019learning, henaff2020data} to overcome CAM's partial activation. \cite{wang2020self} employs pixel-level contrast from positive samples following geometric transformations to extract features known as equivariant features. \cite{ke2021universal} improve WSSS by pixel-to-segment contrast while \cite{yuan2023multi} exploits the similarity and dissimilarity of contrastive sample pairs at the image, region, pixel, and object boundary levels. However, all these methods operate under the assumption that the segments are predetermined or known in advance.

Some others use class-agnostic saliency map that provides rich boundary information in addition to CAMs to enhance the quality of pseudo labels. \cite{joon2017exploiting,zeng2019joint, yao2020saliency} use saliency map either as a part of the pseudo mask or as a saliency feedback for CAM. However, these approaches are sensitive to the errors and noise of the saliency maps \cite{lee2021railroad}. 

Some others focus on learning pairwise semantic affinities to refine the CAM maps. Theoretically, similarities or affinities from images can be learned, and sparse and noisy labels of object regions can be propagated to generate dense and accurate annotations. \cite{ahn2018learning} introduces AffinityNet which
learns the affinities between neighboring pixels from the reliable seeds of the raw CAM maps. It predicts an affinity matrix, facilitating the propagation of CAM maps through a random walk. \cite{xu2023mctformer+} improves the quality of CAMs, and utilizes multiple class tokens to learn class-specific attention maps from transformers rather than CNNs. However, the region labels from the CAM-based methods are noisy and sometimes inaccurate, thereby leading to several challenges in the generation of confident regions. 

\subsection{Foundation model assisted Pseudo-labels}
Lately, there has been a growing trend in leveraging pre-trained large foundation models. \cite{xie2022clims} proposes CLIP-based losses to supervise an auxiliary network for the generation of high-quality CAMs. \cite{lin2023clip} deploys Softmax-GradCAM and class-aware attention-based affinity (CAA) to directly generate CAMs from CLIP. However, CLIP is trained for classification purposes and partial and false activation issues remain with these approaches. Some other works take advantage of the Segment Anything Model (SAM) by \cite{kirillov2023segment} which has trained with over 1 billion masks on 11 million images.  SAM consists of three main components: an image encoder, a prompt encoder, and a mask decoder. Initially, the prompt encoder takes input prompts to guide the mask encoder. Finally, the Mask decoder leverages the image embedding from the image encoder and interactive positional information from the prompt decoder for final pixel label prediction.
In WSSS, \cite{chen2023segment} enhances CAM-based pseudo-labels by including SAM-generated segments based on overlap ratios. \cite{jiang2023segment} improves pseudo-labels by prompting SAM with local maximum points on CAMs. SAM cannot generate class-aware masks, therefore these approaches assign class labels to masks by evaluating their overlap ratio with CAM-based pseudo labels. Despite efforts to mitigate partial activation through the use of SAM, false activation still persists in their approach. This can lead to the erroneous selection of masks. 
Unlike previous methods, to enhance the accuracy of label assignment, we propose leveraging Grounding-DINO, a methodology introduced by \cite{liu2023grounding}, to predict object locations within bounding boxes. Our approach seeks to remedy partial and false activation by searching and hierarchically grouping pseudo-labels within specified regions.

\section{Method}
\begin{figure}
\centering
\includegraphics[width=14cm,height=8cm,keepaspectratio]{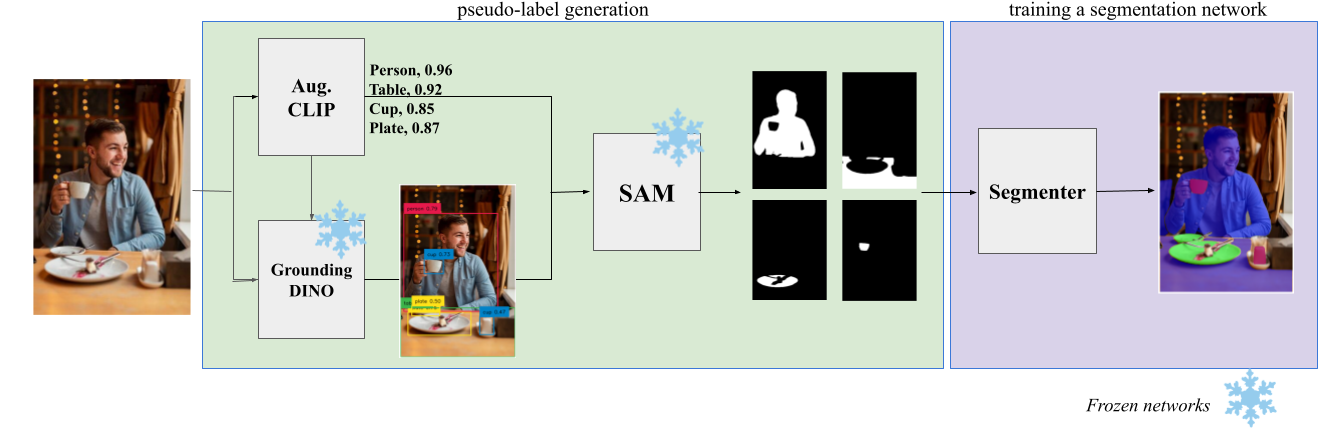}
\caption{\label{fig:pipeline} Overview of the proposed two-stage WSSS training pipeline: (1) Generation of pseudo labels aided by foundation models (green box), and (2) Training of a segmentation network using these pseudo labels (purple box). During the inference phase, the trained  segmenter is used to directly segment the input image. }
\end{figure}

In this section, we present an end-to-end framework to generate high-quality pseudo labels. For each image, an image label
predicted by CLIP and a bounding box containing an object of interest predicted by Grounding-DINO is fed to SAM to generate a class-aware pseudo-label inside the box. Then pairs of images and pseudo-labels are used to train an off-the-shelf segmenter as shown in Figure~\ref{fig:pipeline}.

\subsection{Overview}
Our main idea is to first generate class-aware pseudo-labels for semantic segmentation in an end-to-end manner, which can then be used for supervised learning. As shown in Figure \ref{fig:pipeline}, our proposed framework is composed of a pseudo-label generation module and a training a segmentation module. In the pseudo-label generation module, we leverage the zero-shot transfer capability of SAM to generate pixel-level annotations for images. To get precise masks inside the bounding box and assign correct image labels to the generated masks, two extra guidance prompts need to be provided to SAM: object bounding box and text prompt. We obtain such text and box prompts through CLIP~\cite{radford2021learning} and Grounding-DINO~\cite{liu2023grounding}. First, CLIP is used to retrieve the image labels using a collection of unlabelled training images and a list of class names. Subsequently, the open-set object detector Grounding-DINO takes the image labels as input and predicts corresponding bounding boxes that delineate the referred objects. The text prompt and bounding box information are input to SAM to perform segmentation of the object enclosed within the specified bounding box. The segmentation results from SAM are used as the pseudo labels for the next stage. Subsequently, in the segmentation module, the generated pseudo-labels are used to train an off-the-shelf segmenter.

\subsection{Pseudo-label Generation Module}
The aim of pseudo-label generation module is to predict high-quality class-aware pixel-level annotations for unlabeled images. We achieve this without needing any form of image or pixel-level annotation, relying instead on the semantic categorization ability of SAM inside the bounding box. While SAM is powerful in perceptual grouping, it cannot distinguish between objects, parts, or subparts. We tackle this issue by providing the bounding box information to SAM and defining a hierarchical grouping scheme of masks inside the bounding box. Our framework alleviates the partial and false activation issues caused by CAM-based methods. 
We describe our framework below. 

\paragraph{Language Image Matching } 
In this section, we aim to leverage a language image-matching framework, to predict an image label for each image of interest. To achieve this, a large-scale vision-language model CLIP \cite{radford2021learning} is utilized. CLIP consists of an image encoder and a text encoder, with the primary objective of learning matching embeddings and quantifying the similarity between images and textual content. For any unlabeled image, CLIP is used to predict its corresponding class, selecting the top-n predictions. By employing this approach, an extensive image collection for each class of interest is curated. To improve the performance of CLIP, following the idea of \cite{dong2022clip}, we retrain CLIP's image encoder through \textbf{data augmentation}. 
This data augmentation is conducted as follows: 
(1) First, we initialize the backbone from the CLIP pretraining model. 
(2) Next, we add a new normalization layer for maintaining the stability and consistency of the features across different samples while learning. We also add a fully connected layer that serves as the classification head. After the feature representations are extracted and potentially normalized by the preceding layers (including the LayerNorm layer), they are fed into this fully connected layer. 
(3) Then, we fine-tune CLIP using cross-entropy loss on the augmented dataset. This process enables us to obtain image labels for each image from the finetuned CLIP model. 


\paragraph{Bounding Box Detection}
To predict the bounding boxes of objects of interest, we adopt a promptable object detector, Grounding-DINO~\cite{liu2023grounding}. Grounding-DINO represents a zero-shot object detection framework employing a Swin (Shifted Windows) transformer by \cite{liu2021swin} to extract image features and BERT (Bidirectional Encoder Representations from Transformers) \cite{devlin2018bert} for extracting textual information. Its primary function is to detect objects within an image based on a given text prompt, generating bounding boxes around objects that meet specific text and box thresholds. 
Subsequently, SAM is adopted to generate segmentation masks inside the bounding box. 

\paragraph{Pseudo-label Generation inside the Bounding Box} 
To perform segmentation within the bounding box, the array containing the bounding box and image label information is input to SAM. The steps carried out prior to SAM ensure that SAM possesses the requisite knowledge of the targeted objects, resulting in the generation of accurate class-aware pixel-level labels. While our approach filters the objects outside the bounding box, still many of the masks inside the box are overlapped, covering the whole and part of the objects. To this end, we prioritize the selection of whole masks, over the part masks by computing the overlap ratio between the mask and the bounding box. This prioritization strategy aims to ensure the accurate representation of entire objects before considering their individual parts. Consequently, this can result in the generation of high-quality pseudo-labels. 

\subsection{Segmentation Module}
Upon the generation of high-quality pseudo-labels by SAM, they are utilized as training labels to supervise the training of a segmentation model. 
This involves the employment of any state-of-the-art fully supervised segmentation model. In our experiments, we adopt deeplabV3+ \cite{chen2017deeplab} and Mask2Former \cite{cheng2022masked}, following the approach of other WSSS methods. 
While the training phase involves the above two-stage framework, the inference phase does not rely on running the foundation models but simply utilizes the trained segmentation network to perform the segmentation.

\section{Experiments}
\label{Sec:Experiments}

\subsection{Datasets and Evaluation Metrics}
\label{datasets}
To compare with previous WSSS approaches, we evaluate our model on two different benchmarks: PASCAL VOC 2012 by \cite{everingham2010pascal} and MS COCO 2014 by \cite{lin2014microsoft}. Pascal VOC 2012 consists of 1,464 training images and 1,449 validation images, encompassing 21 categories including a background. MS COCO 2014 contains a total of 82,081 images for training and 40,137 images for validation, featuring 81 object categories including background. For inference, we use the validation split for all datasets. Additionally, for fine-tuning CLIP on the augmented dataset we use Open Image Dataset by \cite{kuznetsova2020open} training set specifically focusing on the subset of categories that present in our target domain, Pascal VOC and MS COCO dataset. We applied category name modifications to the Open Images Dataset to better align with our target datasets. For instance, the category 'motorcycle' was renamed to 'motor bikes,' and 'tv' was adjusted to 'television.' Following common practice by \cite{xie2022clims,yang2024foundation,dong2022clip}, the mean Intersection over Union (mIoU) is adopted as the evaluation metric for all experiments.

\subsection{Implementation details} \label{implementation}
For fine-tuning CLIP, the pre-trained ViT-B/16 \cite{dosovitskiy2020image} image encoder is adopted. 
Similar to \cite{dong2022clip}, we use configuration as Table \ref{finetune config}. For the segmentation network in module two, we employ deeplabv3+ by \cite{chen2018encoder} with ResNet101 backbone architecture by \cite{he2016deep} and Mask2Former by \cite{jain2023oneformer}, with Swin-L backbone architecture by \cite{liu2021swin}. All the training images are resized and center-cropped to 320×320 pixels. While inference, we use the original resolution of images. Other training settings, such as the optimizer, learning rate, etc., are set following \cite{lin2023clip}. Throughout the experiments and ablation study, the IoU threshold is set to 0.3 following the common practice in WSSS.

\begin{table}
  \centering
  \small 
  \begin{tabular}{@{}lcccp{0.9cm}@{}}
    \toprule
    methods & pub./year & backbone & sp. & mIoU\\
    \midrule
    DeepLabV3+ \cite{chen2018encoder} & CVPR18  & R101 & F & 79.5 \\
    Mask2Former \cite{cheng2022masked} & CVPR22 & Swin-L & F & 80.0\\
    \midrule
    CIAN \cite{fan2020cian}& AAAI20 & R101 & I & 64.3\\
    AdvCAM \cite{lee2021anti}& CVPR21 & R101 & I & 67.5\\
    Kweon et al. \cite{kweon2021unlocking}& ICCV21 & WR38 & I & 68.4 \\
    SIPE \cite{chen2022self} & CVPR22 & R101 & I & 68.8\\
    ViT-PCM \cite{rossetti2022max} & ECCV22 & R101 & I & 70.3\\
    CLIMS \cite{xie2022clims} & CVPR22 & R50 & I+C & 70.4\\
    Jiang et al. \cite{jiang2023segment} & arXiv23 & R101 & I+S & 71.1\\
    ToCo \cite{ru2023token} & CVPR23 & WR38 & I & 71.1\\
    MCTformer \cite{xu2022multi} & CVPR22 & WR38 & I & 71.9\\
    WSSS-SAM \cite{chen2023segment} & NeurIPSW23 & R101 & I+S & 72.1\\
    Xu et al. \cite{xu2023learning}& CVPR23 & WR38 & I+C & 72.2 \\
    BECO \cite{rong2023boundary} & CVPR23 & MiT-B2 & I & 73.7\\
    CLIP-ES \cite{lin2023clip} & CVPR23 & R101 & I+C & 73.8 \\
    WeakTr \cite{zhu2023weaktr} & arXiv23 & ViT-S & I & 74.0 \\ 
    
    \midrule
    Ours & PR24 & R101& C+D+S & \textbf{76.9} \\
    Ours & PR24 & Swin-L & C+D+S & \textbf{78.3} \\
    \bottomrule
  \end{tabular}
  \caption{Performance comparison with previous methods of the same setting on the PASCAL VOC 2012 val set. The training supervision type is indicated in the "sp." column, distinguishing between full supervision (F) and image labels (I). Additionally, the inclusion of CLIP (C), Grounding-DINO (D), and SAM (S) is indicated. Also, "pub." refers to the publisher. Best
  WSSS scores are marked in bold.}
  \label{tab:compare pascal}
\end{table}

\begin{table}
  \centering
  \small 
  \begin{tabular}{@{}lcccp{1cm}@{}}
    \toprule
    methods & pub/year & backbone & sp. & mIOU\\
    \midrule
    DeepLabV3+ \cite{chen2018encoder} & CVPR18  & R101 & F & 60.4\\
    Mask2Former \cite{cheng2022masked} & CVPR22 & Swin-L & F & 66.7\\
    \midrule
    Kweon et al. \cite{kweon2021unlocking}& ICCV21 & WR38 & I & 36.4\\
    SIPE \cite{chen2022self} & CVPR22 & R101 & I & 40.6\\
    MCTformer \cite{xu2022multi} & CVPR22 & WR38 & I & 42.0\\
    ToCo \cite{ru2023token} & CVPR23 & WR38 & I & 42.3\\
    AdvCAM \cite{lee2021anti}& CVPR21 & R101 & I & 44.4\\ 
    ViT-PCM \cite{rossetti2022max} & ECCV22 & R101 & I & 45.0\\
    BECO \cite{rong2023boundary} & CVPR23 & MiT-B2 & I & 45.1\\
    CLIP-ES \cite{lin2023clip} & CVPR23 & R101 & I+C & 45.4 \\
    Xu et al. \cite{xu2023learning}& CVPR23 & WR38 & I+C & 45.9 \\
    WeakTr \cite{zhu2023weaktr} & arXiv23 & ViT-S & I & 46.9 \\ 
    WSSS-SAM \cite{chen2023segment} & NeurIPSW23 & R101 & I+S & 45.9\\
    
    \midrule
    Ours & PR24 & R101& C+D+S & \textbf{48.5} \\
    Ours & PR24 & Swin-L& C+D+S & \textbf{49.9} \\
    \bottomrule
  \end{tabular}
  \caption{Performance comparison with previous methods of the the same setting on the MS COCO 2014 val set. The training supervision type is indicated in the "sp." column, distinguishing between full supervision (F) and image labels (I). Additionally, the inclusion of CLIP (C), Grounding-DINO (D), and SAM (S) is indicated. Also, "pub." refers to the publisher. Best
  WSSS scores are marked in bold.}
  \label{tab:compare coco}
\end{table}

\begin{table}[htbp]
  \centering
  \small
  \begin{tabular}{@{}lccc}
    \toprule
    methods & pseudo-label & mIOU\\
    \midrule
    Jiang et al. \cite{jiang2023segment} &  SAM & 61.9\\
    AdvCAM \cite{lee2021anti} & RW+CRF  & 60.4\\
    MCTformer \cite{xu2022multi} &  RW+CRF &66.7\\
    CLIMS \cite{xie2022clims}&  RW+CRF &36.4\\
    ViT-PCM \cite{rossetti2022max} & CRF & 40.6\\
    MCTformer \cite{xu2022multi} & RW+CRF & 42.0\\
    CLIP-ES \cite{lin2023clip} & CAA+CRF & 42.3\\
    WeakTr \cite{zhu2023weaktr} & CRF & 68.7 \\ 
    WSSS-SAM \cite{chen2023segment} & CAA+SAM &  79.6\\
    \midrule
    Ours & SAM & \textbf{87.2} \\
    \bottomrule
  \end{tabular}
  \caption{Pseudo-label quality on PASCAL VOC 2012 training set with the generation techniques. The "pseudo-label" refers to the pseudo-label generation or enhancement method including training affinity networks (RW), dense CRF (CRF), class-aware attention-based affinity (CAA), and SAM. The best results are marked in bold.} 
  \label{tab:compare pseudo}
\end{table}

\begin{table}[htbp]
  \centering
  \small
  \begin{tabular}{p{4cm}@{}p{2.5cm}@{}}
    \toprule
    configuration & value \\
    \midrule
    optimizer & AdamW\\
    learning rate & $2 \times 10^{-5}$\\
    weight decay & 0.7 \\
    batch size & 32\\
    learning rate schedule & cosine decay \\
    epochs & 50\\
    \bottomrule
  \end{tabular}
  \caption{\label{finetune config}Configuration for augmneted CLIP.}
\end{table}

\subsection{Comparison to the State-of-the-Art} 
\paragraph{Quantitative Results}
\label{quantitative} 
To evaluate the effectiveness of our approach, we compare
our final segmentation result with the state-of-the-art WSSS methods on both PASCAL VOC and MS COCO datasets. The results are demonstrated in Table \ref{tab:compare pascal} and \ref{tab:compare coco}. On both PASCAL VOC and MS COCO, our approach outperforms others by a significant margin. It’s worth mentioning that our method compared to all previous approaches, does not use image-level labels, alleviating the manually annotating burden of large-scale datasets. Furthermore, we compare the quality of our generated pseudo-labels with previous approaches on PASCAL VOC in Table \ref{tab:compare pseudo}. The results show that our approach generates more accurate pseudo labels than all previous methods by a substantial margin. Compared to two previous approaches, WSSS-SAM \cite{chen2023segment} and \cite{jiang2023segment} that use SAM in the pseudo-label generation
module, our framework boosts the quality of labels to 9.55\% and 40.87\%. The reason is that WSSS-SAM \cite{chen2023segment} utilizes SAM for enhancing CAM-based pseudo-labels which may fail if the classifier either activates on incorrect objects or fails to activate on the target objects. Also, \cite{jiang2023segment} directly uses SAM for generating pseudo labels; however, SAM cannot perform well when box prompts are not provided. Our approach is able to generate high-quality pseudo-labels by adopting SAM to perform inside the bounding box which leads to significant performance improvement in final segmentation results.

\begin{figure}[htbp]
\small
\centering
\includegraphics[width=19cm,height=5.8cm,keepaspectratio]{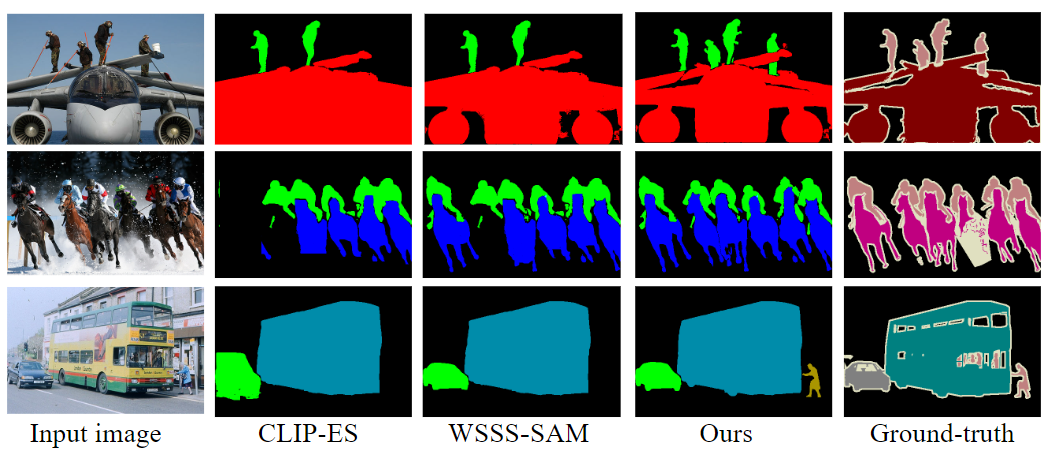}
\caption{\label{sam} The pseudo-labels generated by our proposed framework and comparison between WSSS-SAM and CLIP-ES on PASCAL VOC 2012 training set.}
\end{figure}

\begin{figure}[htbp]
\centering
\includegraphics[width=20cm,height=14cm,keepaspectratio]{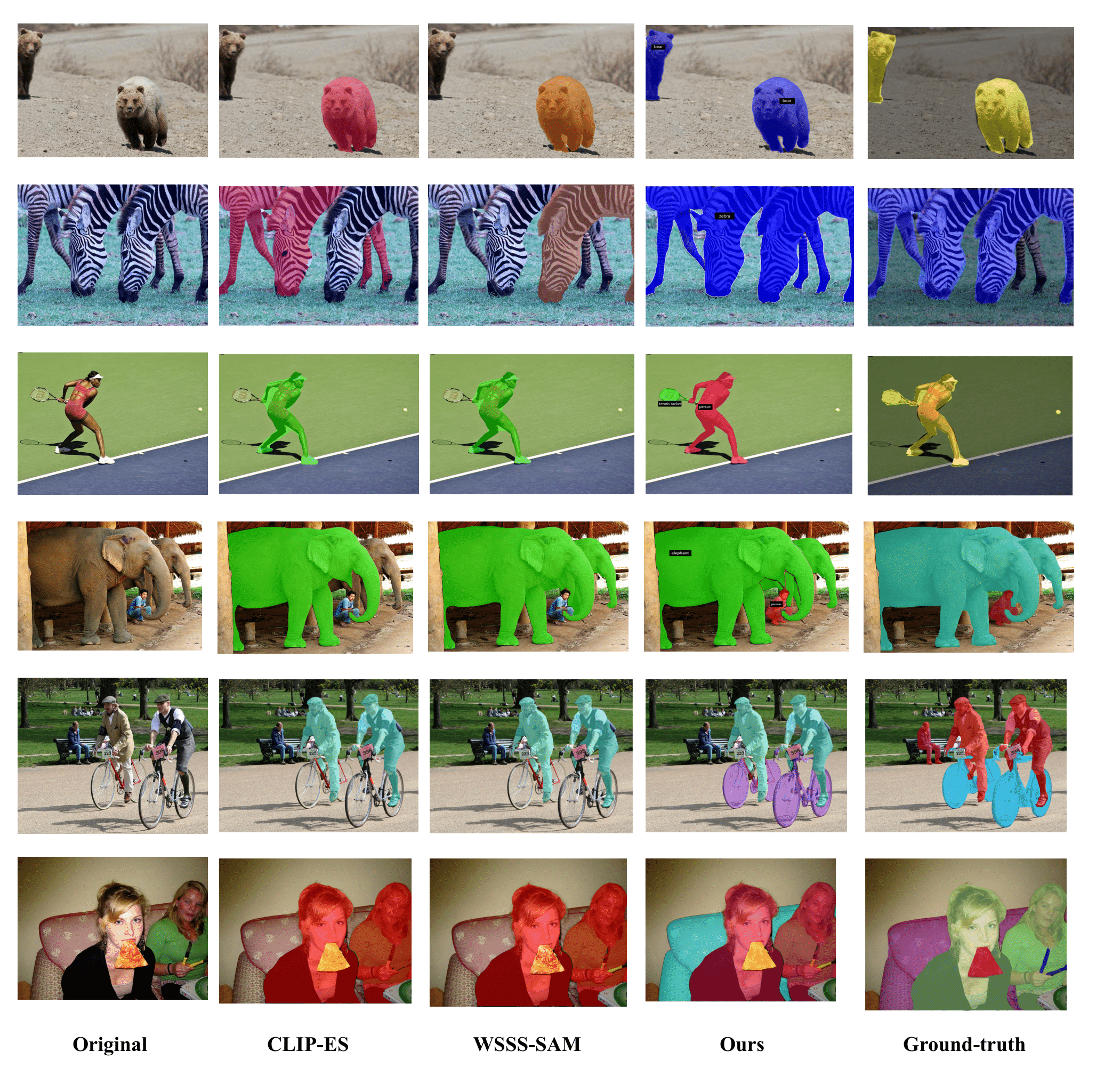}
\caption{\label{fig:fig4} Visualization of the final segmentation results, CLIP-ES, and WSSS-SAM on MS COCO 2014 validation set. }
\end{figure}

\paragraph{Qualitative Results}
\label{qualitative}
We visually compared our predicted pseudo-labels with previous models in Figure \ref{sam}. Our pseudo-label generation module segments more complete regions and precise boundaries for objects compared to WSSS-SAM \cite{chen2023segment} and CLIP-ES \cite{lin2023clip}. CLIP-ES suffers from partial activation and false activation of CAMs. For example in the second row, CAM is not activated for all horses in the image due to partial activation; also the space between the legs of the horses is activated to the false activation. While WSSS-SAM leverages SAM to alleviate the issues with CAM, the visual results demonstrate that the problem still remains due to the inactivation or false activation of CAMs and the inefficiency of the voting scheme between SAM pseudo-labels and CAM-based seeds.
Figure \ref{sam} demonstrates the superiority of using SAM-based pseudo-label generation approaches to CAM-based methods in a WSSS setting.   

Figure \ref{fig:fig4} compares our final segmentation results with WSSS-SAM and CLIP-ES.  These results demonstrate the superiority of our proposed framework in segmenting whole object regions. Furthermore, our model performs better at segmenting fine-grained details particularly the boundaries within the images such as
the legs of the bear and zebras in the first and second rows, and the left arm of the person in the third row.

\subsection{Ablation Studies}
\label{ablation}

To certify the effectiveness of our design, we present a series of experiments on PASCAL VOC 2012 val set. Our final segmentation network is Mask2Former trained with pseudo-labels.
\paragraph{The analysis of the proposed components}
We evaluate the performance of each component of our pseudo-label generation module (w/o aug.) in Table \ref{tab:component}. In addition, we analyze the performance of each component after augmented CLIP (w/ aug.) on Open Image Dataset. As shown, augmentation improves the performance of CLIP, Grounding-DINO and SAM to 6.68\%, 4.87\%, and 3.93\%. 

\begin{table}
  \centering
  \small
  \caption{Comparing the performance of each pseudo-label generation module component on PASCAL VOC 2012 val set in \% before and after augmentation. "cls.", "bx." and "pseul." refer to the classifier, bounding box detector, and pseudo-label generator. Also, "perf.", "w/ aug." and "w/o aug." indicate performance, with and without augmentation, respectively.} 
  \label{tab:component}
  \begin{tabular}{@{}ll|ll}
    \toprule
    component & perf. & component & perf.\\
    \midrule
    cls. (w/o aug.) &  mAP: 90.1 & cls. (w/ aug.) &  mAP: 96.2\\
    bx.  (w/o aug.) & mAP: 88.0 & bx.  (w/ aug.) & mAP: 92.3\\
    pseul. (w/o aug.) &  mIOU: 83.9 &  pseul. (w/ aug.) &  mIOU: 87.2\\
    \bottomrule
  \end{tabular}
\end{table}

\paragraph{Enhancing the performance by using supervision}
We investigate the effect of using different kinds of supervision including image and box labels. In the first experiment, Grounding-DINO takes the image labels and generates box prompts. All other components will remain the same. In the second experiment, we eliminate CLIP and Grounding-DINO and feed SAM with image and box labels. Our experiments demonstrate that by using image labels the quality of pseudo-labels and final segmentation results improve by +1.26 and +1.4, while using image and box labels enhances the performance by +4.9 and +3.7 in the cost of manually annotating the dataset. 

\begin{table}
  \centering
  \small
  \caption{Improving the performance of the pseudo label generation module and the final segmenter while using any forming of supervision including image labels and box labels on PASCAL VOC 2012 val set. Compare our performance with previous methods in Table \ref{tab:compare pascal}.} 
  \label{tab:subs}
  \begin{tabular}{@{}cccc}
    \toprule
    image labels & box labels & pseudo-label generator & final segmenter\\
    \midrule
    $\times$ &  $\times$ & 87.2 & 78.3\\
    \checkmark &  $\times$ & 88.3 & 79.4\\
    \checkmark &  \checkmark & 91.5 & 81.2\\
    \bottomrule
  \end{tabular}
\end{table}

\section{Conclusion}
This paper proposes a pioneering framework by applying SAM in the bounding box in 
WSSS setting. We address two main issues with CAM-based methods: partial and false activation. By using SAM inside the bounding box and defining a hierarchical grouping scheme, we guide our model to select masks that are within a box area around the object ( to avoid false activation) and we choose the whole objects over the parts (to avoid the partial activation). We also eliminate image-level labels and outperform previous approaches.
Our pseudo-label generation module is capable of producing fine-grained labels that are both class-aware and object-aware. Our approach shows consistent improvement over the SOTA WSSS methods on both PASCAL VOC and MS-COCO datasets. 

\paragraph{Data availability statement} All data supporting the findings of this study are available within the paper.

\section*{Declarations}
\textbf{Competing interests} We do not have any conflict of interest related to the manuscript.








\bibliographystyle{elsarticle-num}
\bibliography{egbib} 

\begin{thebibliography}{10}
\expandafter\ifx\csname url\endcsname\relax
  \def\url#1{\texttt{#1}}\fi
\expandafter\ifx\csname urlprefix\endcsname\relax\def\urlprefix{URL }\fi
\expandafter\ifx\csname href\endcsname\relax
  \def\href#1#2{#2} \def\path#1{#1}\fi

\bibitem{wang2023cut}
X.~Wang, R.~Girdhar, S.~X. Yu, I.~Misra, Cut and learn for unsupervised object detection and instance segmentation, in: Proceedings of the IEEE/CVF Conference on Computer Vision and Pattern Recognition, 2023, pp. 3124--3134.

\bibitem{lee2021anti}
J.~Lee, E.~Kim, S.~Yoon, Anti-adversarially manipulated attributions for weakly and semi-supervised semantic segmentation, in: Proceedings of the IEEE/CVF Conference on Computer Vision and Pattern Recognition, 2021, pp. 4071--4080.

\bibitem{wei2017object}
Y.~Wei, J.~Feng, X.~Liang, M.-M. Cheng, Y.~Zhao, S.~Yan, Object region mining with adversarial erasing: A simple classification to semantic segmentation approach, in: Proceedings of the IEEE conference on computer vision and pattern recognition, 2017, pp. 1568--1576.

\bibitem{khoreva2017simple}
A.~Khoreva, R.~Benenson, J.~Hosang, M.~Hein, B.~Schiele, Simple does it: Weakly supervised instance and semantic segmentation, in: Proceedings of the IEEE conference on computer vision and pattern recognition, 2017, pp. 876--885.

\bibitem{lee2021bbam}
J.~Lee, J.~Yi, C.~Shin, S.~Yoon, Bbam: Bounding box attribution map for weakly supervised semantic and instance segmentation, in: Proceedings of the IEEE/CVF conference on computer vision and pattern recognition, 2021, pp. 2643--2652.

\bibitem{oh2021background}
Y.~Oh, B.~Kim, B.~Ham, Background-aware pooling and noise-aware loss for weakly-supervised semantic segmentation, in: Proceedings of the IEEE/CVF conference on computer vision and pattern recognition, 2021, pp. 6913--6922.

\bibitem{li2023transcam}
R.~Li, Z.~Mai, Z.~Zhang, J.~Jang, S.~Sanner, Transcam: Transformer attention-based cam refinement for weakly supervised semantic segmentation, Journal of Visual Communication and Image Representation 92 (2023) 103800.

\bibitem{yin2022transfgu}
Z.~Yin, P.~Wang, F.~Wang, X.~Xu, H.~Zhang, H.~Li, R.~Jin, Transfgu: a top-down approach to fine-grained unsupervised semantic segmentation, in: European conference on computer vision, Springer, 2022, pp. 73--89.

\bibitem{lee2021railroad}
S.~Lee, M.~Lee, J.~Lee, H.~Shim, Railroad is not a train: Saliency as pseudo-pixel supervision for weakly supervised semantic segmentation, in: Proceedings of the IEEE/CVF conference on computer vision and pattern recognition, 2021, pp. 5495--5505.

\bibitem{ahn2018learning}
J.~Ahn, S.~Kwak, Learning pixel-level semantic affinity with image-level supervision for weakly supervised semantic segmentation, in: Proceedings of the IEEE conference on computer vision and pattern recognition, 2018, pp. 4981--4990.

\bibitem{xie2022clims}
J.~Xie, X.~Hou, K.~Ye, L.~Shen, Clims: Cross language image matching for weakly supervised semantic segmentation, in: Proceedings of the IEEE/CVF Conference on Computer Vision and Pattern Recognition, 2022, pp. 4483--4492.

\bibitem{kirillov2023segment}
A.~Kirillov, E.~Mintun, N.~Ravi, H.~Mao, C.~Rolland, L.~Gustafson, T.~Xiao, S.~Whitehead, A.~C. Berg, W.-Y. Lo, et~al., arXiv preprint arXiv:2304.02643 (2023).

\bibitem{chen2023weakly}
Z.~Chen, Q.~Sun, Weakly-supervised semantic segmentation with image-level labels: from traditional models to foundation models, arXiv preprint arXiv:2310.13026 (2023).

\bibitem{yang2024foundation}
X.~Yang, X.~Gong, Foundation model assisted weakly supervised semantic segmentation, in: Proceedings of the IEEE/CVF Winter Conference on Applications of Computer Vision, 2024, pp. 523--532.

\bibitem{liu2023grounding}
S.~Liu, Z.~Zeng, T.~Ren, F.~Li, H.~Zhang, J.~Yang, C.~Li, J.~Yang, H.~Su, J.~Zhu, et~al., Grounding dino: Marrying dino with grounded pre-training for open-set object detection, arXiv preprint arXiv:2303.05499 (2023).

\bibitem{radford2021learning}
A.~Radford, J.~W. Kim, C.~Hallacy, A.~Ramesh, G.~Goh, S.~Agarwal, G.~Sastry, A.~Askell, P.~Mishkin, J.~Clark, et~al., Learning transferable visual models from natural language supervision, in: International conference on machine learning, PMLR, 2021, pp. 8748--8763.

\bibitem{araslanov2020single}
N.~Araslanov, S.~Roth, Single-stage semantic segmentation from image labels, in: Proceedings of the IEEE/CVF Conference on Computer Vision and Pattern Recognition, 2020, pp. 4253--4262.

\bibitem{zhang2022end}
B.~Zhang, J.~Xiao, Y.~Wei, K.~Huang, S.~Luo, Y.~Zhao, End-to-end weakly supervised semantic segmentation with reliable region mining, Pattern Recognition 128 (2022) 108663.

\bibitem{yuan2023multi}
K.~Yuan, G.~Schaefer, Y.-K. Lai, Y.~Wang, X.~Liu, L.~Guan, H.~Fang, A multi-strategy contrastive learning framework for weakly supervised semantic segmentation, Pattern Recognition 137 (2023) 109298.

\bibitem{yi2022weakly}
S.~Yi, H.~Ma, X.~Wang, T.~Hu, X.~Li, Y.~Wang, Weakly-supervised semantic segmentation with superpixel guided local and global consistency, Pattern Recognition 124 (2022) 108504.

\bibitem{kho2022exploiting}
S.~Kho, P.~Lee, W.~Lee, M.~Ki, H.~Byun, Exploiting shape cues for weakly supervised semantic segmentation, Pattern Recognition 132 (2022) 108953.

\bibitem{luo2021weakly}
W.~Luo, M.~Yang, W.~Zheng, Weakly-supervised semantic segmentation with saliency and incremental supervision updating, Pattern Recognition 115 (2021) 107858.

\bibitem{zhou2016learning}
B.~Zhou, A.~Khosla, A.~Lapedriza, A.~Oliva, A.~Torralba, Learning deep features for discriminative localization, in: Proceedings of the IEEE conference on computer vision and pattern recognition, 2016, pp. 2921--2929.

\bibitem{chen2023segment}
T.~Chen, Z.~Mai, R.~Li, W.-l. Chao, Segment anything model (sam) enhanced pseudo labels for weakly supervised semantic segmentation, arXiv preprint arXiv:2305.05803 (2023).

\bibitem{kumar2017hide}
K.~Kumar~Singh, Y.~Jae~Lee, Hide-and-seek: Forcing a network to be meticulous for weakly-supervised object and action localization, in: Proceedings of the IEEE International Conference on Computer Vision, 2017, pp. 3524--3533.

\bibitem{chang2020weakly}
Y.-T. Chang, Q.~Wang, W.-C. Hung, R.~Piramuthu, Y.-H. Tsai, M.-H. Yang, Weakly-supervised semantic segmentation via sub-category exploration, in: Proceedings of the IEEE/CVF Conference on Computer Vision and Pattern Recognition, 2020, pp. 8991--9000.

\bibitem{zhang2018adversarial}
X.~Zhang, Y.~Wei, J.~Feng, Y.~Yang, T.~S. Huang, Adversarial complementary learning for weakly supervised object localization, in: Proceedings of the IEEE conference on computer vision and pattern recognition, 2018, pp. 1325--1334.

\bibitem{li2018tell}
K.~Li, Z.~Wu, K.-C. Peng, J.~Ernst, Y.~Fu, Tell me where to look: Guided attention inference network, in: Proceedings of the IEEE conference on computer vision and pattern recognition, 2018, pp. 9215--9223.

\bibitem{hou2018self}
Q.~Hou, P.~Jiang, Y.~Wei, M.-M. Cheng, Self-erasing network for integral object attention, Advances in Neural Information Processing Systems 31 (2018).

\bibitem{lee2019ficklenet}
J.~Lee, E.~Kim, S.~Lee, J.~Lee, S.~Yoon, Ficklenet: Weakly and semi-supervised semantic image segmentation using stochastic inference, in: Proceedings of the IEEE/CVF Conference on Computer Vision and Pattern Recognition, 2019, pp. 5267--5276.

\bibitem{bachman2019learning}
P.~Bachman, R.~D. Hjelm, W.~Buchwalter, Learning representations by maximizing mutual information across views, Advances in neural information processing systems 32 (2019).

\bibitem{henaff2020data}
O.~Henaff, Data-efficient image recognition with contrastive predictive coding, in: International conference on machine learning, PMLR, 2020, pp. 4182--4192.

\bibitem{wang2020self}
Y.~Wang, J.~Zhang, M.~Kan, S.~Shan, X.~Chen, Self-supervised equivariant attention mechanism for weakly supervised semantic segmentation, in: Proceedings of the IEEE/CVF conference on computer vision and pattern recognition, 2020, pp. 12275--12284.

\bibitem{ke2021universal}
T.-W. Ke, J.-J. Hwang, S.~X. Yu, Universal weakly supervised segmentation by pixel-to-segment contrastive learning, arXiv preprint arXiv:2105.00957 (2021).

\bibitem{joon2017exploiting}
S.~Joon~Oh, R.~Benenson, A.~Khoreva, Z.~Akata, M.~Fritz, B.~Schiele, Exploiting saliency for object segmentation from image level labels, in: Proceedings of the IEEE conference on computer vision and pattern recognition, 2017, pp. 4410--4419.

\bibitem{zeng2019joint}
Y.~Zeng, Y.~Zhuge, H.~Lu, L.~Zhang, Joint learning of saliency detection and weakly supervised semantic segmentation, in: Proceedings of the IEEE/CVF international conference on computer vision, 2019, pp. 7223--7233.

\bibitem{yao2020saliency}
Q.~Yao, X.~Gong, Saliency guided self-attention network for weakly and semi-supervised semantic segmentation, IEEE Access 8 (2020) 14413--14423.

\bibitem{xu2023mctformer+}
L.~Xu, M.~Bennamoun, F.~Boussaid, H.~Laga, W.~Ouyang, D.~Xu, Mctformer+: Multi-class token transformer for weakly supervised semantic segmentation, arXiv preprint arXiv:2308.03005 (2023).

\bibitem{lin2023clip}
Y.~Lin, M.~Chen, W.~Wang, B.~Wu, K.~Li, B.~Lin, H.~Liu, X.~He, Clip is also an efficient segmenter: A text-driven approach for weakly supervised semantic segmentation, in: Proceedings of the IEEE/CVF Conference on Computer Vision and Pattern Recognition, 2023, pp. 15305--15314.

\bibitem{jiang2023segment}
P.-T. Jiang, Y.~Yang, Segment anything is a good pseudo-label generator for weakly supervised semantic segmentation, arXiv preprint arXiv:2305.01275 (2023).

\bibitem{dong2022clip}
X.~Dong, J.~Bao, T.~Zhang, D.~Chen, S.~Gu, W.~Zhang, L.~Yuan, D.~Chen, F.~Wen, N.~Yu, Clip itself is a strong fine-tuner: Achieving 85.7\% and 88.0\% top-1 accuracy with vit-b and vit-l on imagenet, arXiv preprint arXiv:2212.06138 (2022).

\bibitem{liu2021swin}
Z.~Liu, Y.~Lin, Y.~Cao, H.~Hu, Y.~Wei, Z.~Zhang, S.~Lin, B.~Guo, Swin transformer: Hierarchical vision transformer using shifted windows, in: Proceedings of the IEEE/CVF international conference on computer vision, 2021, pp. 10012--10022.

\bibitem{devlin2018bert}
J.~Devlin, M.-W. Chang, K.~Lee, K.~Toutanova, Bert: Pre-training of deep bidirectional transformers for language understanding, arXiv preprint arXiv:1810.04805 (2018).

\bibitem{chen2017deeplab}
L.-C. Chen, G.~Papandreou, I.~Kokkinos, K.~Murphy, A.~L. Yuille, Deeplab: Semantic image segmentation with deep convolutional nets, atrous convolution, and fully connected crfs, IEEE transactions on pattern analysis and machine intelligence 40~(4) (2017) 834--848.

\bibitem{cheng2022masked}
B.~Cheng, I.~Misra, A.~G. Schwing, A.~Kirillov, R.~Girdhar, Masked-attention mask transformer for universal image segmentation, in: Proceedings of the IEEE/CVF conference on computer vision and pattern recognition, 2022, pp. 1290--1299.

\bibitem{everingham2010pascal}
M.~Everingham, L.~Van~Gool, C.~K. Williams, J.~Winn, A.~Zisserman, The pascal visual object classes (voc) challenge, International journal of computer vision 88 (2010) 303--338.

\bibitem{lin2014microsoft}
T.-Y. Lin, M.~Maire, S.~Belongie, J.~Hays, P.~Perona, D.~Ramanan, P.~Doll{\'a}r, C.~L. Zitnick, Microsoft coco: Common objects in context, in: Computer Vision--ECCV 2014: 13th European Conference, Zurich, Switzerland, September 6-12, 2014, Proceedings, Part V 13, Springer, 2014, pp. 740--755.

\bibitem{kuznetsova2020open}
A.~Kuznetsova, H.~Rom, N.~Alldrin, J.~Uijlings, I.~Krasin, J.~Pont-Tuset, S.~Kamali, S.~Popov, M.~Malloci, A.~Kolesnikov, et~al., The open images dataset v4: Unified image classification, object detection, and visual relationship detection at scale, International Journal of Computer Vision 128~(7) (2020) 1956--1981.

\bibitem{dosovitskiy2020image}
A.~Dosovitskiy, L.~Beyer, A.~Kolesnikov, D.~Weissenborn, X.~Zhai, T.~Unterthiner, M.~Dehghani, M.~Minderer, G.~Heigold, S.~Gelly, et~al., An image is worth 16x16 words: Transformers for image recognition at scale, arXiv preprint arXiv:2010.11929 (2020).

\bibitem{chen2018encoder}
L.-C. Chen, Y.~Zhu, G.~Papandreou, F.~Schroff, H.~Adam, Encoder-decoder with atrous separable convolution for semantic image segmentation, in: Proceedings of the European conference on computer vision (ECCV), 2018, pp. 801--818.

\bibitem{he2016deep}
K.~He, X.~Zhang, S.~Ren, J.~Sun, Deep residual learning for image recognition, in: Proceedings of the IEEE conference on computer vision and pattern recognition, 2016, pp. 770--778.

\bibitem{jain2023oneformer}
J.~Jain, J.~Li, M.~T. Chiu, A.~Hassani, N.~Orlov, H.~Shi, Oneformer: One transformer to rule universal image segmentation, in: Proceedings of the IEEE/CVF Conference on Computer Vision and Pattern Recognition, 2023, pp. 2989--2998.

\bibitem{fan2020cian}
J.~Fan, Z.~Zhang, T.~Tan, C.~Song, J.~Xiao, Cian: Cross-image affinity net for weakly supervised semantic segmentation, in: Proceedings of the AAAI Conference on Artificial Intelligence, Vol.~34, 2020, pp. 10762--10769.

\bibitem{kweon2021unlocking}
H.~Kweon, S.-H. Yoon, H.~Kim, D.~Park, K.-J. Yoon, Unlocking the potential of ordinary classifier: Class-specific adversarial erasing framework for weakly supervised semantic segmentation, in: Proceedings of the IEEE/CVF international conference on computer vision, 2021, pp. 6994--7003.

\bibitem{chen2022self}
Q.~Chen, L.~Yang, J.-H. Lai, X.~Xie, Self-supervised image-specific prototype exploration for weakly supervised semantic segmentation, in: Proceedings of the IEEE/CVF Conference on Computer Vision and Pattern Recognition, 2022, pp. 4288--4298.

\bibitem{rossetti2022max}
S.~Rossetti, D.~Zappia, M.~Sanzari, M.~Schaerf, F.~Pirri, Max pooling with vision transformers reconciles class and shape in weakly supervised semantic segmentation, in: European Conference on Computer Vision, Springer, 2022, pp. 446--463.

\bibitem{ru2023token}
L.~Ru, H.~Zheng, Y.~Zhan, B.~Du, Token contrast for weakly-supervised semantic segmentation, in: Proceedings of the IEEE/CVF Conference on Computer Vision and Pattern Recognition, 2023, pp. 3093--3102.

\bibitem{xu2022multi}
L.~Xu, W.~Ouyang, M.~Bennamoun, F.~Boussaid, D.~Xu, Multi-class token transformer for weakly supervised semantic segmentation, in: Proceedings of the IEEE/CVF Conference on Computer Vision and Pattern Recognition, 2022, pp. 4310--4319.

\bibitem{xu2023learning}
L.~Xu, W.~Ouyang, M.~Bennamoun, F.~Boussaid, D.~Xu, Learning multi-modal class-specific tokens for weakly supervised dense object localization, in: Proceedings of the IEEE/CVF Conference on Computer Vision and Pattern Recognition, 2023, pp. 19596--19605.

\bibitem{rong2023boundary}
S.~Rong, B.~Tu, Z.~Wang, J.~Li, Boundary-enhanced co-training for weakly supervised semantic segmentation, in: Proceedings of the IEEE/CVF Conference on Computer Vision and Pattern Recognition, 2023, pp. 19574--19584.

\bibitem{zhu2023weaktr}
L.~Zhu, Y.~Li, J.~Fang, Y.~Liu, H.~Xin, W.~Liu, X.~Wang, Weaktr: Exploring plain vision transformer for weakly-supervised semantic segmentation, arXiv preprint arXiv:2304.01184 (2023).

\end{thebibliography}

\end{document}